\documentclass{egpubl}
\usepackage{pg2024}
 
\SpecialIssuePaper         

\CGFStandardLicense

\usepackage[T1]{fontenc}
\usepackage{dfadobe}  

\usepackage{cite}  
\BibtexOrBiblatex
\electronicVersion
\PrintedOrElectronic
\ifpdf \usepackage[pdftex]{graphicx} \pdfcompresslevel=9
\else \usepackage[dvips]{graphicx} \fi

\usepackage{egweblnk} 

\usepackage{xcolor}
\usepackage{amsmath}
\usepackage{amssymb}
\usepackage{hyperref}       
\usepackage{cleveref}

\title[GSEditPro]%
      {GSEditPro: 3D Gaussian Splatting Editing with Attention-based Progressive Localization}

\author[Y.\ Sun \& R. Tian \& X.\ Han \& X.\ Liu \& Y.\ Zhang \& K.\ Xu]
{\parbox{\textwidth}{\centering 
Y.\ Sun\thanks{Equal contributions to this work.}$^{1}$\orcid{0009-0008-1406-9099},\
    R. Tian\footnotemark[1]$^{1}$\orcid{0009-0000-8807-4802},\
    X.\ Han\footnotemark[1]$^{1}$\orcid{0009-0006-6218-9170},\
    X.\ Liu$^{2}$\orcid{0009-0009-1836-4017},\
    Y.\ Zhang\thanks{Corresponding author.}$^{1}$\orcid{0000-0002-9621-7321}\ and
    K.\ Xu$^{2}$\orcid{0000-0002-9054-0216}
        }
        \\
{\parbox{\textwidth}{\centering $^1$State Key Laboratory for Novel Software Technology of Nanjing\ University, China\\
         $^2$National\ University\ of\ Defense\ Technology, China
       }
}
}

\begin{document}

\teaser{
 \includegraphics[width=0.65\linewidth]{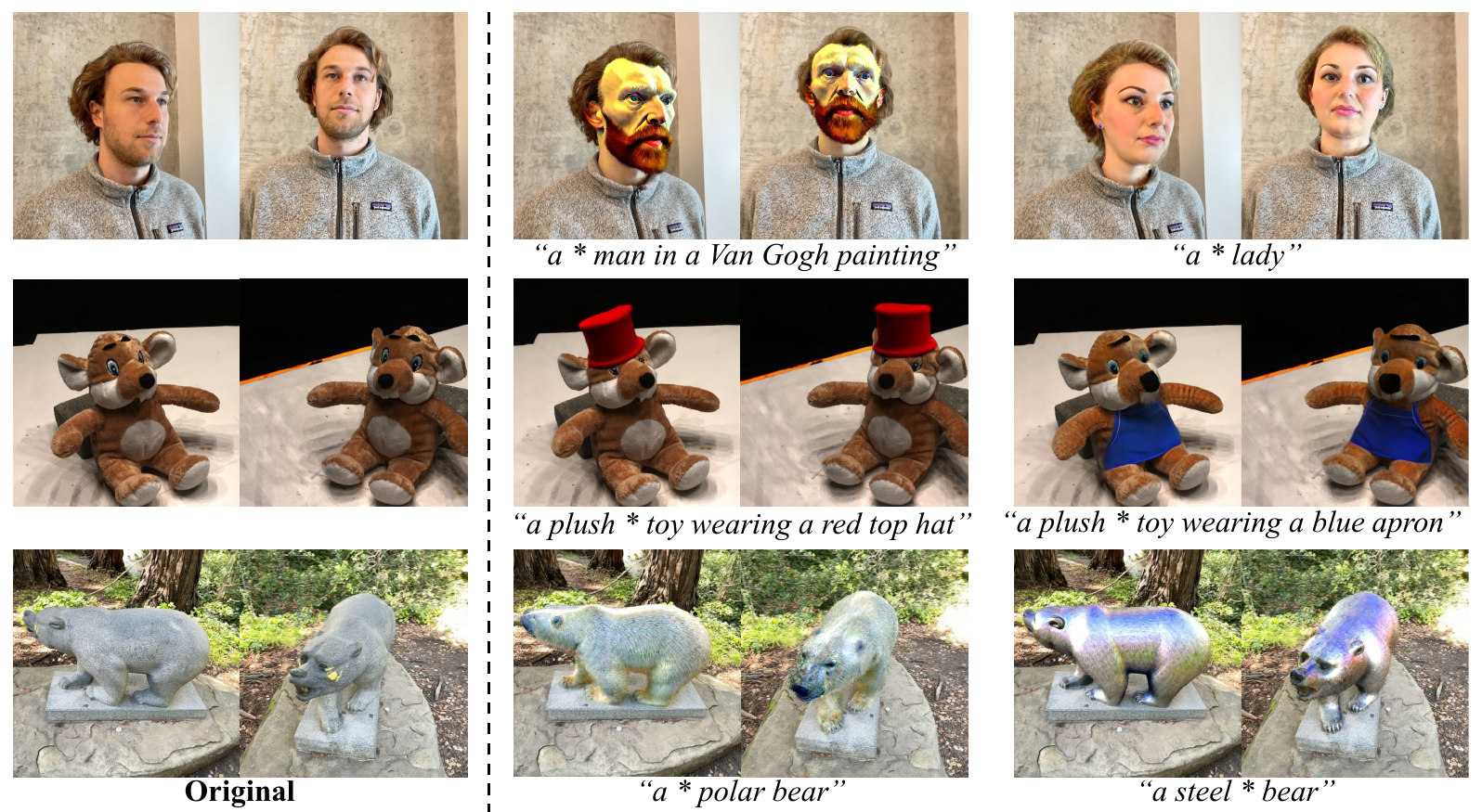}
 \centering
  \caption{\textbf{Results of GSEditPro.} GSEditPro enables users to conduct high-quality editing in various scenes using text prompts only.}
\label{fig:teaser}
}
\maketitle
\begin{abstract}
    With the emergence of large-scale Text-to-Image(T2I) models and implicit 3D representations like Neural Radiance Fields (NeRF), many text-driven generative editing methods based on NeRF have appeared. However, the implicit encoding of geometric and textural information poses challenges in accurately locating and controlling objects during editing. Recently, significant advancements have been made in the editing methods of 3D Gaussian Splatting, a real-time rendering technology that relies on explicit representation. However, these methods still suffer from issues including inaccurate localization and limited manipulation over editing. To tackle these challenges, we propose GSEditPro, a novel 3D scene editing framework which allows users to perform various creative and precise editing using text prompts only. Leveraging the explicit nature of the 3D Gaussian distribution, we introduce an attention-based progressive localization module to add semantic labels to each Gaussian during rendering. This enables precise localization on editing areas by classifying Gaussians based on their relevance to the editing prompts derived from cross-attention layers of the T2I model. Furthermore, we present an innovative editing optimization method based on 3D Gaussian Splatting, obtaining stable and refined editing results through the guidance of Score Distillation Sampling and pseudo ground truth. We prove the efficacy of our method through extensive experiments. \\

\end{abstract}  
\section{Introduction}
In the rapidly evolving field of computer graphics, developing user-friendly methods for 3D generation and editing is crucial, as these methods can be widely applied in domains such as virtual reality and digital gaming.

In the field of 3D generation, text-based model generation technology~\cite{Poole2022DreamFusionTU,Lin2022Magic3DHT,metzer2023latent,wang2024prolificdreamer} has made significant progress due to the success of large-scale Text-to-Image (T2I) models~\cite{saharia2022photorealistic,ramesh2022hierarchical,yu2022scaling}. These models demonstrate remarkable creativity and significantly reduce the cost of model generation, gaining increasing attention. 
However, they often lack editing abilities, and even slight text prompt variations may lead to different output results.
In addition to 3D generation tasks, editing existing 3D models is also crucial, which enables efficient and precise modifications to 3D models, thereby increasing the flexibility and adaptability of the existing models for various applications.

In recent years, the emergence of implicit 3D representation Neural Radiance Fields (NeRF)\cite{mildenhall2021nerf} has made significant progress in scene reconstruction and novel view synthesis. The high-fidelity rendering ability of NeRF and its significant scalability provide excellent support for subsequent work.
Consequently, most text-driven 3D editing techniques\cite{haque2023instruct,mikaeili2023sked,wang2022clip} have been designed based on NeRF for quite some time.
However, editing neural radiance fields is difficult due to its implicit encoding of shape and texture information in high-dimensional neural network features. 
Thus, accurate locating and direct modification during the editing process are challenging, hindering the obtainment of precise and high-quality editing results, thereby impeding their practical applications.
A pioneering work recently emerging in the field of 3d reconstruction is 3D Gaussian Splatting (3D-GS)\cite{kerbl20233d}, a real-time rendering technology based on explicit representation. The explicit nature of 3D-GS gives it a significant advantage in editing tasks. Each 3D Gaussian distribution exists independently, allowing for editing 3D scenes easily by directly manipulating the 3D Gaussians required for editing constraints. 
Recently, some editing methods\cite{chen2023gaussianeditor,fang2023gaussianeditor,ye2023gaussian,zhuang2024tip} based on 3D-GS have emerged. However, they still encounter various issues such as inaccurate locating or requiring users to manually locate editing areas in some cases\cite{zhuang2024tip, chen2023gaussianeditor}, difficulty in ensuring consistency of non-editing areas before and after editing\cite{ye2023gaussian, chen2023gaussianeditor}, inability to perform object insertion operations effectively\cite{chen2023gaussianeditor, fang2023gaussianeditor}, and failure to guarantee consistency between different viewpoints after editing\cite{fang2023gaussianeditor}.

To overcome these issues, we propose a novel text-driven editing framework based on 3D-GS called GSEditPro, which enables users to perform 3D editing intuitively and precisely using text prompts. Our framework achieves this through two key designs: (1) Attention-based editing area localization in 3D: We leverage the explicit representation advantage of 3D-GS to classify Gaussians based on their relevance to the attention maps derived from cross-attention layers of T2I models, assigning semantic labels to each Gaussian, thereby obtaining accurate 3D editing mask areas. (2) Guidance for a detailed Optimization from DreamBooth and pseudo-GT images: We create an optimization process that balances generative capability and detail preservation. It uses simple text prompts to effectively perform 3D scene editing by conducting score distillation sampling within the 3D editing mask area, thus ensuring high-quality editing. Additionally, we maintain the details of the scenes by constructing pseudo-GT images to ensure consistency of the irrelevant regions using pixel-level guidance.

We conducted experiments using the proposed method in various synthetic and real-world 3D scenes. The experiments demonstrate that our editing method can achieve precise editing both on object changing and object insertion with irrelevant areas naturally preserved after editing. Furthermore, since editing is accomplished through simple text prompts, our method is highly user-friendly, showcasing significant practical application potential. Qualitative and quantitative comparisons also indicate that our method outperforms previous methods in terms of editing accuracy, visual fidelity, and user satisfaction.

Our contributions can be summarized as follows:

1. We propose GSEditPro, a novel 3D editing method that enables users to perform various creative and precise editing operations using only text prompts. This approach is more convenient than previous Gaussian editing methods, which require additional user input as prior. Our extensive experiments demonstrate that our framework still offers advantages in both qualitative and quantitative metrics with user-friendly interactions.

2. We design a method to add semantic labels to Gaussians using the cross-attention mechanism of the T2I model and the explicit representation advantage of 3D-GS. Our special attention-based localization module assists in achieving more accurate 3D editing area localization and more effective editing control.

3. We present how to preserve details with pixel-level guidance, which creates a pseudo-GT image using our localization module to minimize unnecessary modifications and guide 3D Gaussian rendering for more detailed results. 

\section{Related Work}
\subsection{Text-guided Image Generation and Editing}

Today, numerous methods have attained impressive outcomes in the realm of text-driven image generation and editing. T2I diffusion models\cite {ho2022cascaded,rombach2022high,saharia2022photorealistic} based on large-scale image-text data demonstrate diverse and high-quality image generation capabilities that align well with text prompts. However, these models do not offer the ability to modify the generated images. Prompt-to-Prompt\cite{hertz2022prompt} utilizes images provided by users to generate images based on text prompts while editing them simultaneously, providing an intuitive editing method. DreamBooth\cite{ruiz2023dreambooth} adjusts diffusion models using an L2 reconstruction loss and proposes a preservation loss to avoid overfitting. It obtains extensive updated model parameters, providing the capability to generate high-quality images and perform editing. 
Additionally, some methods\cite{wang2024instancediffusion,zhang2023adding}incorporate spatial conditions that control object positions during the generation process. This enhances the model's ability to handle various input conditions, allowing for finer control over the generated image results. These methods have achieved excellent results in the 2D domain, but extending them directly from 2D to 3D is not straightforward.

\subsection{Text-to-3D}

With the development of T2I generation models, interest in the Text-to-3D domain is continuously increasing. However, directly applying diffusion models in the 3D domain is a challenging task since it requires keeping the consistency of different views. DreamField\cite{jain2022zero} employs the image-text embedding model CLIP\cite{radford2021learning} to guide the optimization of NeRF\cite{mildenhall2021nerf}, successfully generating 3D shapes from text prompts. DreamFusion\cite{Poole2022DreamFusionTU} first proposed the score distillation sampling (SDS) loss, which obtains priors from a pre-trained T2I model and optimizes the neural radiance field during training. Based on DreamFusion, a series of works~\cite{Lin2022Magic3DHT,metzer2023latent,raj2023dreambooth3d} adopted a similar optimization process and achieved better generation results by refining the process and employing different SDS guidance methods. Furthermore, recent works ~\cite{zhou2024gala3d,di2024hyper,li2024dreamscene}utilize 3D-GS\cite{kerbl20233d} as their 3D representation, enabling rapid generation of 3D models based on text prompts. However, the generation results of these methods are easily influenced by the effectiveness of text prompts, and they are limited to generating 3D models, unable to edit existing 3D scenes.

\subsection{Text-guided 3D Editing}

Text-guided neural radiance field editing has gained significant attention as a new research area. EditNeRF\cite{liu2021editing} pioneered this field, offering the ability to edit both the shape and color of NeRF based on implicit encoding. Subsequently, some methods\cite{mikaeili2023sked,wang2022clip} began combining NeRF\cite{mildenhall2021nerf} and diffusion models. For instance, Instruct-NeRF2NeRF\cite{haque2023instruct} utilized a text-based diffusion model\cite{brooks2023instructpix2pix}) to modify rendered images based on the user’s instructions and gradually modify the neural radiance field, achieving excellent editing effects. However, due to the implicit representation of NeRF, these editing methods lack precise control over editing regions. Therefore, previous methods\cite{zhuang2023dreameditor,sella2023vox} adopted explicit representation methods such as grids and voxels to improve the quality of local editing. However, these methods have not obtained satisfactory editing results for real-world scenes. Concurrent work ConsistentDreamer\cite{chen2024consistdreamer} adds 3D-consistent structured noise to rendered multi-view images, and applies self-supervised consistency training using consistency-warped images, generating 3D consistently edited images from 2D diffusion models.

The introduction of 3D-GS\cite{kerbl20233d} presents an opportunity to overcome this limitation. Its explicit representation enhances control over editing regions. GaussianEditor\cite{chen2023gaussianeditor} utilizes 3D-GS as scene representation and employs semantic tracking technology to track the cloning and splitting of Gaussians, resulting in more accurate scene editing results. Another GaussianEditor\cite{fang2023gaussianeditor} leverages large language models to extract Regions of Interest (RoI) from text prompts and converts them to the image space using segmentation models. Then it trains them using the loss proposed in SA3D\cite{cen2023segment} to elevate RoI to the 3D scene. GaussianGrouping\cite{ye2023gaussian} adds identity encoding to Gaussians, classifying them and providing the ability to modify 3D objects. However, the editing results of these methods lack consistency across different viewpoints, and they fail to add objects to the scene or request additional input from users, which results in significant inconvenience. Concurrent works aim to address this problem. DGE\cite{chen2024dge} injects features from selected key views into the diffusion network through correspondence matching with epipolar constraints, enabling direct editing views of a 3D model with a text-based image editor. From the perspective of image editing, VCEdit\cite{wang2024view} intergrates two view-consistent modules into the Gaussian editing framework. TIGER\cite{xu2024tiger} introduces Coherent Score Distillation, which combines a 2D image editing diffusion model with a multi-view diffusion model for score distillation, resulting in multi-view consistent outcomes.
\section{Our Method}
\begin{figure*}[ht]
  \centering
  \includegraphics[width=0.9\linewidth]{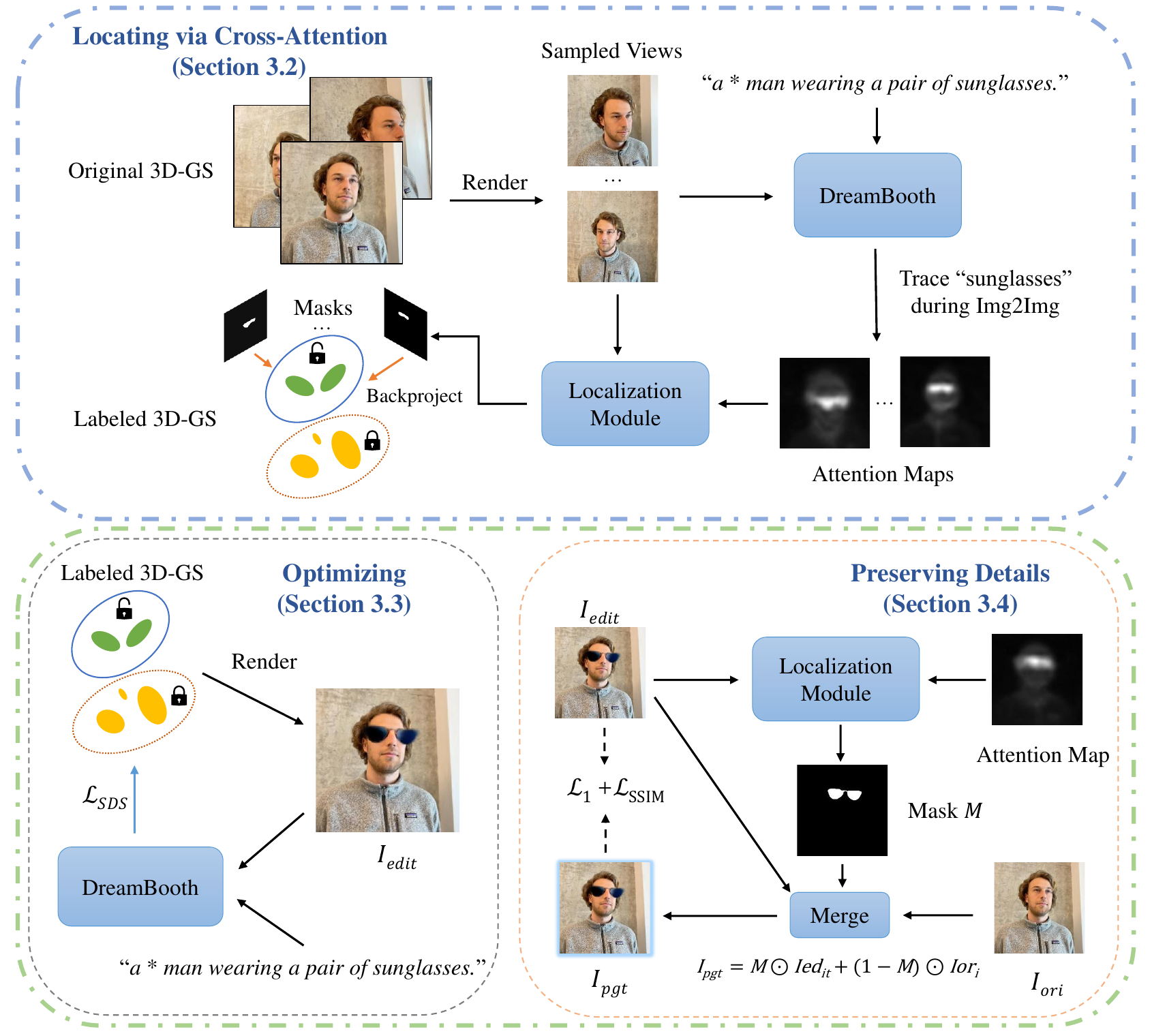}
  \caption{\label{fig:ex3}%
           \textbf{Method Overview.} GSEditPro edits scenes represented by 3D Gaussian Splatting using text prompt only. The key process of our method contains three parts: 1) Locating Gaussian editing regions via cross-attention, which assigns semantic labels to each Gaussian and determines whether the gradients can be propagated between them (Section~\ref{sec:locating}); 2) Optimizing editing regions using DreamBooth, which uses $\mathcal{L}_{SDS}$ as the guidance to optimize Gaussian parameters iteratively (Section~\ref{sec:Optim}); 3) Preserving details with pixel-level guidance, which creates a pseudo-GT image to guide 3D Gaussian rendering for more detailed results (Section~\ref{sec:Preservation}).}
\end{figure*}
GSEditPro is a novel editing framework designed to edit a pre-trained 3D-GS scene according to the provided text prompt. It alters the geometry and appearance of the objects of interest within the original scene while ensuring that the 3D content unrelated to the prompt remains unchanged.

The overall framework of GSEditPro, as illustrated in Figure~\ref{fig:ex3}, consists of two main stages. Firstly, we design an attention-based localization module that employs a T2I model and the cross-attention mechanism to locate the editing region in the 3D space using the keywords in the text prompt, as elaborated in Section \ref{sec:locating}. 
Secondly, building upon 3D Gaussians, we implement scene editing leveraging Score Distillation Sampling (SDS) loss with Dreambooth\cite{ruiz2023dreambooth}, as detailed in Sections \ref{sec:Optim}. 
We employ the attention-based localization module at the pixel level to further preserve the unrelated areas during the editing process, as depicted in Section \ref{sec:Preservation}. 
By integrating optimization and progressive localization, our method achieves precise and detailed local editing.

\subsection{Preliminary}
\label{sec:Pre}
\textbf{3D Gaussian Splatting.} 3D Gaussian Splatting (3D-GS) \cite{kerbl20233d} is an explicit representation method for 3D scenes, utilizing a set of anisotropic 3D Gaussians to represent the scene, denoted as $G={g_1,g_2,...,g_\mathrm{N}}$, where $g_\mathrm{i}=\{\mu,\Sigma,c,\alpha\}$, $i\in{1,...,N}$. Among them, $\mu$ denotes the center position of the Gaussian, $\Sigma$ represents the 3D covariance matrix, $c$ is the RGB color represented by spherical harmonic coefficients, and $\alpha$ denotes opacity. 3D-GS employs a differentiable splatting rendering method, enabling high-quality real-time rendering. The splatting rendering process can be formulated as:
\begin{equation}
    C=\sum_\mathrm{i\in\mathcal{N}}c_\mathrm{i}\sigma_\mathrm{i}\prod_\mathrm{j=1}^\mathrm{i-1}(1-\sigma_\mathrm{j})
\end{equation}
where $\mathcal{N}$ represents the number of Gaussians contributing to the ray, $\sigma_\mathrm{i}=\alpha_\mathrm{i}e^\mathrm{-\frac{1}{2}(x_i)^{T}\Sigma^{-1}(x_i)}$ denotes the influence of the Gaussian on the image pixel, and $x_\mathrm{i}$ is the distance between the pixel and the center of the $i$-th Gaussian.

\textbf{SDS Loss.} DreamFusion\cite{Poole2022DreamFusionTU} first introduced the Score Distillation Sampling (SDS) loss, which guides the generation of 3D models by extracting prior knowledge from the T2I diffusion model. It first adds noise at level $t$ to a randomly rendered view $I$ to obtain $I_\mathrm{t}$. Then it uses a pre-trained diffusion model $\Phi$ to predict the added noise under the condition of $I_\mathrm{t}$ and the text prompt $y$. The SDS loss is calculated as the gradient for each pixel as follows:
\begin{equation}
\nabla_{\mathrm{\theta}}\mathcal{L}_{\mathrm{SDS}}(\Phi,\mathit{I}=g(\theta)) = \mathbb{E}_{\mathrm{\epsilon,t}}[w(t)(\epsilon_{\mathrm{\Phi}}(\mathit{I}_\mathrm{t}:y,t)-\epsilon)\frac{\partial I}{\partial \theta}]
\end{equation}
where $w(t)$ is a weighting function based on the noise level $t$, $\theta$ represents the parameters of the neural radiance field, and $g$ denotes the rendering process function. During training, gradients are back-propagated to $\theta$, guiding the rendering results of the neural radiance field to be more similar to the images generated by the T2I diffusion model based on text prompts.

\subsection{Locating Gaussian Editing Regions via Cross-Attention}
\label{sec:locating}
Initially, we locate the region of interest for modification within 3D space based on text prompts, laying the foundation for the editing framework.
Previous 3D editing approaches\cite{li2023climatenerf,kobayashi2022decomposing} have commonly relied on static 2D masks for determining the editing region, yet such methods suffer from limitations in ensuring consistency of the edited scene across views. With dynamic changes occurring in the 3D representation during training, these masks can become inaccurate or even ineffective. Additionally, some works\cite{xu2022deforming,yuan2022nerf} have utilized 3D masks for locating. Still, they need manual operation to get the 3D masks in some cases, which is conducted with complex rules and may result in the imprecision of the located region.

The cross-attention layers inside the T2I diffusion model can capture the relationship between the generated image and each word \cite{hertz2022prompt}. Similarly, during our editing process, we need to manipulate the target objects within the 3D Gaussians under the control of text prompts through the T2I diffusion model. Therefore, we propose an attention-based localization module that utilizes the 2D probability maps generated by the cross-attention layers as masks for each view, determining which regions need editing in 2D views. These 2D maps are then processed as point prompts for the large-scale segmentation model Segment Anything Model(SAM)\cite{kirillov2023segment} to obtain a more precise mask for the target region. After that, we backproject the 2D masks into 3D space and mark the Gaussians that need editing, enabling precise localization of the editing region explicitly in Gaussians before training.

Concretely, we sample rendering output in various views using COLMAP\cite{schonberger2016structure} cameras and fine-tune the Stable Diffusion\cite{rombach2022high} using DreamBooth\cite{ruiz2023dreambooth}. DreamBooth is a method that fine-tunes the large-scale text-to-image (T2I) model around a specific target subject, denoted as "*" or other symbols, to ensure its ability to generate images similar to the input data. To strengthen the generating stability and ability of the fine-tuned diffusion model, we set the class prompt as the target editing prompt. The preservation loss of DreamBooth will encourage the diffusion model to treat this special class as the default generating style, which increases the accuracy of attention maps as well. 

Furthermore, we collect the attention maps of the target words during the Img2ImgPipe of DreamBooth, which generates several images based on our editing prompt. These maps from cross-attention layers represent the rough or possible position of the editing area depending on whether it exists in the original scene, which means our method can have a reasonable localization of incorporation editing with the prior of the Diffusion model. Our localization module will decide how to locate the region according to the existence of the target object. Suppose the text prompt for editing is about adding new objects to the scene. In that case, the localization module chooses the filtered attention maps directly as the preliminary results, and the threshold is set as 0.5 in our experiments. Considering maps lack precision, preliminary results are clustered using the clustering algorithm DBSCAN\cite{ester1996density} to filter out outliers further to get the final 2D masks. When editing existing objects in the scene, our module first tries to use Language-based SAM\cite{kirillov2023segment} to segment them in sampled views. However, the results of the SAM based on the language prompt differ from the views which will result in bad results of the masks. And it always fails to segment the part of the target editing objects. So we will improve the results when they have a small overlap over the attention maps. The traced maps are filtered and then clustered as mentioned before. The localization module chooses points of the processed maps as point prompts for the SAM, with the top 5 points selected based on the highest attention map values as positive ones, while the negative point prompts are chosen based on the lowest 3 values. After that SAM will segment a precise mask of the target for each view. Masks are back-projected during the differentiable rending process similar to GaussianEditor\cite{chen2023gaussianeditor} and we only allow gradients to propagate within the labeled Gaussians whose weights of back-projection bigger than the threshold. Finally, our method finishes Locating Gaussian editing regions explicitly and assigns the Gaussians their binary labels in 3D.

\subsection{Optimizing Editing Regions using DreamBooth}
\label{sec:Optim}
After locating the editing regions, we propose an optimization scheme for 3D Gaussian editing. To achieve text-based 3D editing, we introduce the diffusion model in the optimization stage. 
After training on our target dataset, DreamBooth\cite{ruiz2023dreambooth} possesses sufficient generation ability to guide the training of 3D Gaussians. We utilize the SDS loss proposed by DreamFusion\cite{Poole2022DreamFusionTU} as the guiding loss function. After obtaining the prompt for editing and the images rendered from random views during training, they are collectively used as inputs to compute $\mathcal{L}_\mathrm{SDS}$ in DreamBooth. This loss is then employed during the back-propagation process to guide the cloning and splitting of the Gaussians, as well as the changes in their parameters. The computation can be formulated as follows:
\begin{equation}
  \mathcal{L}_{SDS} = w(t)(\epsilon_\Phi(x_\mathrm{t},t,T)-\epsilon)^2
  \label{eq:5}
\end{equation}
where \textit{w(t)} is the weight of SDS decided by timestep $t$, $\epsilon_\Phi$ is the denoiser of the diffusion model to compute the noise which will be removed, $x_\mathrm{t}$ is the embedding vector of the noised image, \textit{T} is the text prompt input. To conveniently control all losses in the method with weight, $\mathcal{L}_\mathrm{SDS}$ adopts the squared error between real noise and predicted noise.

In Section~\ref{sec:locating}, we have already finished locating editing regions and only allowed the gradients to be backpropagated between the Gaussians to be edited. Therefore, during each training iteration, $\mathcal{L}_{SDS}$ serves as a 2D guidance to optimize Gaussian parameters iteratively. This process matches the rendering results with the text guidance of the editing, obtaining desired editing results after sufficient training.

Inspired by GaussianEditor\cite{chen2023gaussianeditor}, the parameters of the GS model are supposed to be constrained according to its existing generations, which will prevent it from being exposed to the randomness of the loss function. And it is represented as:
\begin{equation}
    L_\mathrm{anchor}^\mathrm{P} = \sum_\mathrm{i=0}^\mathrm{n} \lambda_\mathrm{i} (P_\mathrm{i} - \hat{P}_\mathrm{i})^2
    \label{eq:6}
\end{equation}
where $P$ denotes the property of Gaussians including the position $x$, the scaling $s$, the rotation $q$, the transparent $\alpha$ and the color $c$. And n is the maximum generation now. $\hat{P}$ represents the saved anchor state. $\lambda_\mathrm{i}$ refers to the strength of the loss applied, which will grow with the increase of the number of terms.
 \begin{figure}[htb]
  \centering
  \includegraphics[width=1.0\linewidth]{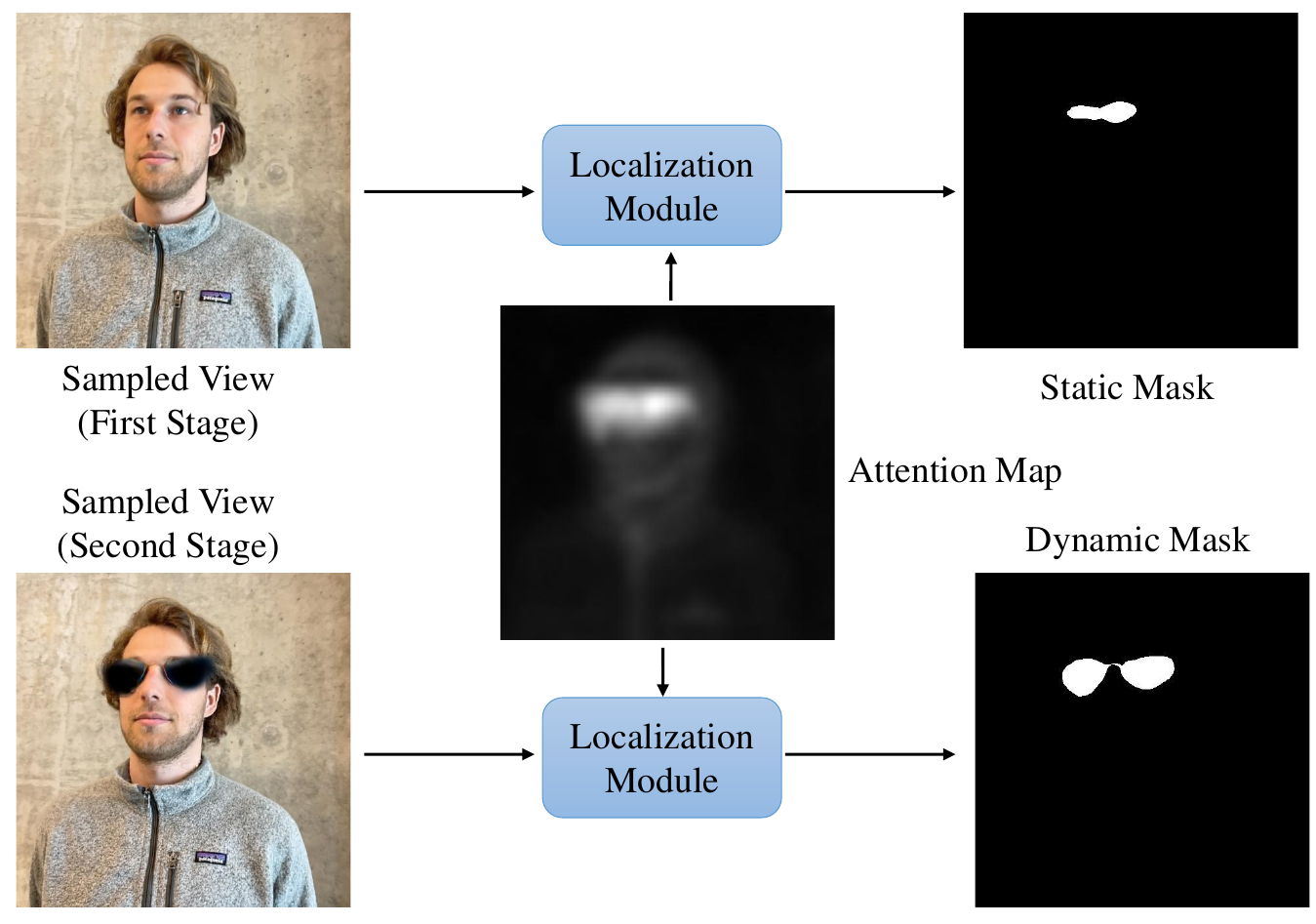}
  \caption{\label{fig:ex2}
           \textbf{The visual difference of the Mask between the two stages.} The static Mask is a guessed sketch from the DreamBooth, and the dynamic mask is located in real-time after 3D-GS capable of rendering decent edited results.}
\end{figure}
\subsection{Preserving Details with Pixel-Level Guidance}
\label{sec:Preservation}

The strong fluidity of Gaussians makes it easy to cause editing beyond the desired region. We propose a method to maintain the accuracy of the editing region at the pixel level. As mentioned in Section~\ref{sec:locating}, our approach has already locked the Gaussian regions to be edited after locating, with gradients only passing through the target Gaussians. However, due to the strong fluidity of Gaussians, there may still be changes outside the editing region in rendered results. We create a pseudo-GT image for each rendered image during training to ensure consistency between the editing results and the original images. The pseudo-GT image is generated by combining the modified parts in the current rendered result with the unedited parts of the initial rendered result as shown in the bottom right of Figure~\ref{fig:ex3}. Note that in our overview the pseudo-GT image is almost the same as the rendered image because they are similar to each other indeed when our method converges. We then use $\mathcal{L}_{1}$ and $\mathcal{L}_\mathrm{D-SSIM}$ losses to constrain the similarity between the current rendered result and the pseudo-GT image, ensuring that after back-propagation, the overall result shifts towards greater consistency.

In order to get a pseudo-GT image to guide 3D Gaussian rendering for desired editing results, we need an appropriate mask to separate the editing region from others accurately. We divide the generation of this mask into two stages as shown in Figure~\ref{fig:ex2}, with the main difference lying in the method of selecting masks.

In the first stage, as shown in Section~\ref{sec:locating}, a mask suitable for locating the editing region can be obtained for each rendered view through the localization module. Therefore, during the initial 2000 iterations of rendering, we utilize this static mask to construct a coarse pseudo-GT image.

In the second stage, when the editing results of the Gaussian rendering have a roughly formed shape, we reuse the localization module introduced in Section~\ref{sec:locating} to locate a dynamic mask for generating a more reliable pseudo-GT image, which changes dynamically during the training optimization. The calculation method for obtaining the pseudo-GT image through masking is as follows:
\begin{equation}
  I_\mathrm{pgt} = M \odot I_\mathrm{edit} + (1 - M) \odot I_\mathrm{ori}
  \label{eq:7}
\end{equation}
where $I_\mathrm{pgt}$ is the pseudo-GT image, $M$ is the obtained mask, $I_\mathrm{edit}$ is the edited image, and $I_\mathrm{ori}$ is the original image. After obtaining $I_\mathrm{pgt}$, we calculate the pixel-wise loss with the edited image as follows:
\begin{equation}
  \mathcal{L}_\mathrm{Preservation} = \lambda_\mathrm{l1} \mathcal{L}_1(I_\mathrm{edit},I_\mathrm{pgt}) + \lambda_\mathrm{ssim} \mathcal{L}_\mathrm{D-SSIM}(I_\mathrm{edit},I_\mathrm{pgt})
  \label{eq:8}
\end{equation}
where $\lambda_\mathrm{l1}$ and $\lambda_\mathrm{ssim}$ represent the weighting values assigned to $\mathcal{L}_1$ and $\mathcal{L}_\mathrm{D-SSIM}$ respectively. Experiment results indicate that our pixel-wise editing consistency preservation method achieves optimal results at this stage. For specific details, please refer to the ablation experiments.

Thus, the loss function $\mathcal{L}$ of our method can be expressed as:
\begin{equation}
  \mathcal{L} = \lambda_\mathrm{sds}\mathcal{L}_\mathrm{SDS} + \mathcal{L}_\mathrm{Preservation} + \sum_\mathrm{P \in \{x, s, q, \alpha, c\}}{\lambda_\mathrm{P} L_\mathrm{anchor}^\mathrm{P}}
  \label{eq:9}
\end{equation}
where $\lambda_\mathrm{sds}$ is the weighting value assigned by $\mathcal{L}_\mathrm{SDS}$ defined in  Equation~\ref{eq:5} and $\lambda_\mathrm{P}$ is the weighting value for different anchor loss of the Gaussian parameters $x, s, q, \alpha, c$.
\begin{figure*}[ht]
  \centering
  \includegraphics[width=1\linewidth]{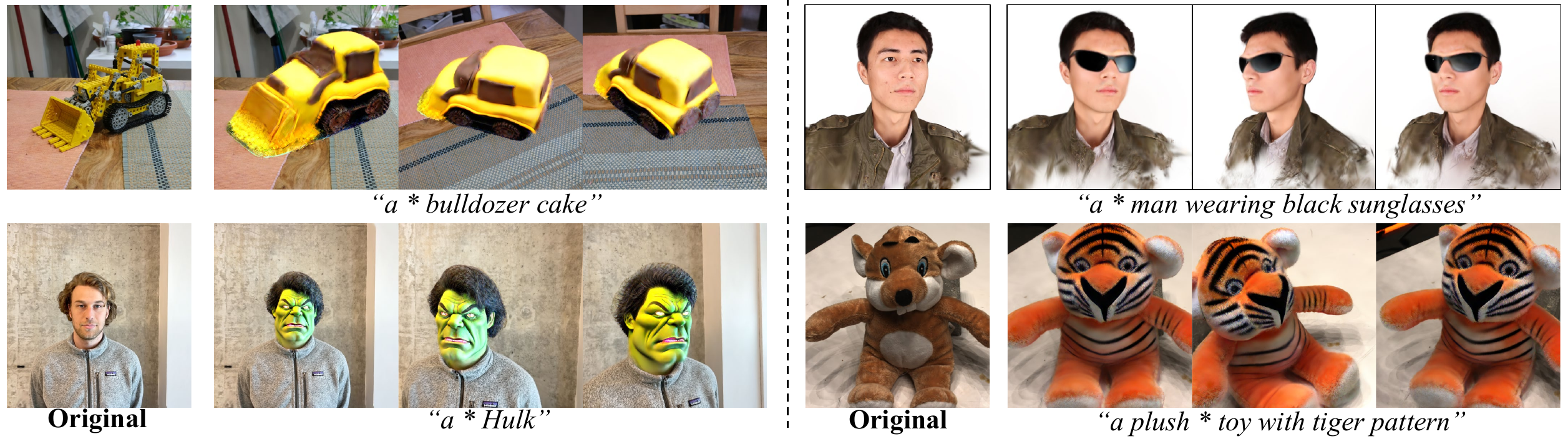}
  \caption{\label{fig:more_result}
        \textbf{More multi-view results of our method in different scenes.} The left column is the original view and the other three columns are the multi-view editing results. Our method is capable of conducting various kinds of editing in different scenes. Different from most previous methods, our method succeeds in altering the geometry and appearance of objects detailedly.
   }
\end{figure*}
\begin{figure*}[htp]
  \centering
  \includegraphics[width=0.95\linewidth]{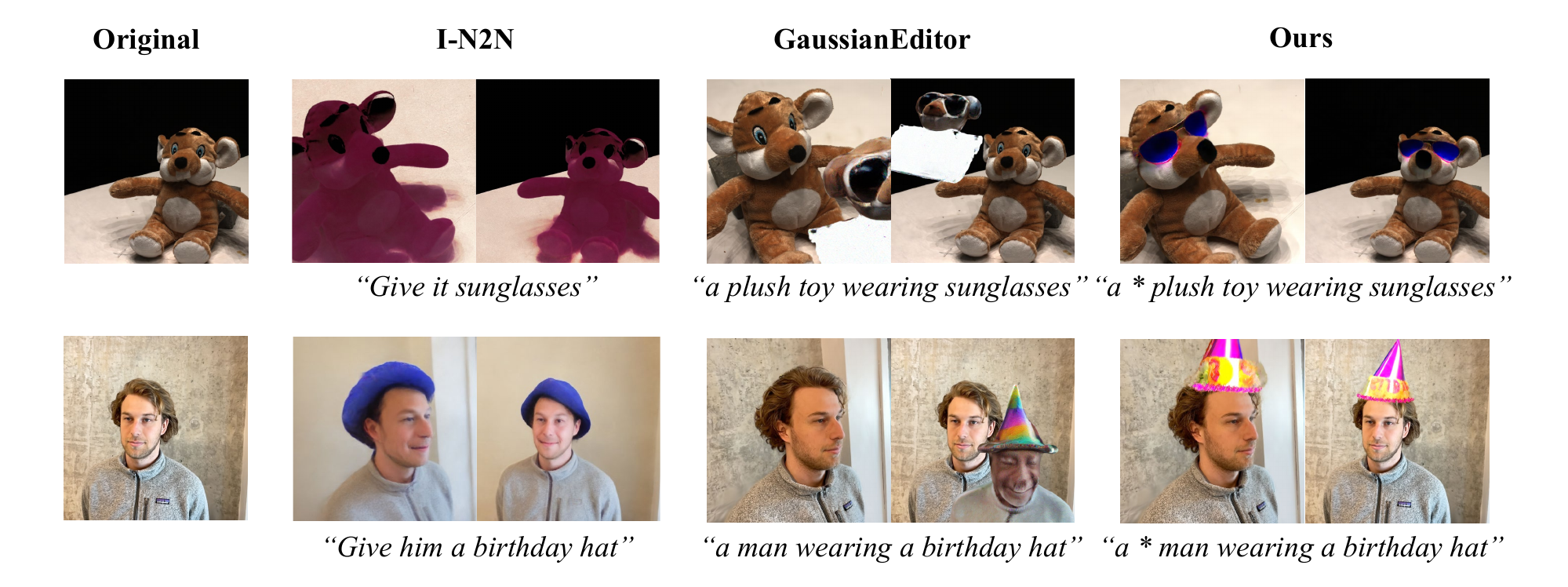}
  \caption{\label{fig:comp-add}
        \textbf{Comparisons on adding objects to the given scene.} Results of the GaussianEditor shown above are generated without manually adjusting the estimated depth during training. Our method gives attention to proper locations and generates satisfactory results. In contrast, GaussianEditor\cite{chen2023gaussianeditor} incorrectly positions Gaussians, and I-N2N\cite{haque2023instruct} fails to edit as instructed. 
   }
\end{figure*}

\begin{figure*}[tbp]
  \centering
  \includegraphics[width=0.85\linewidth]{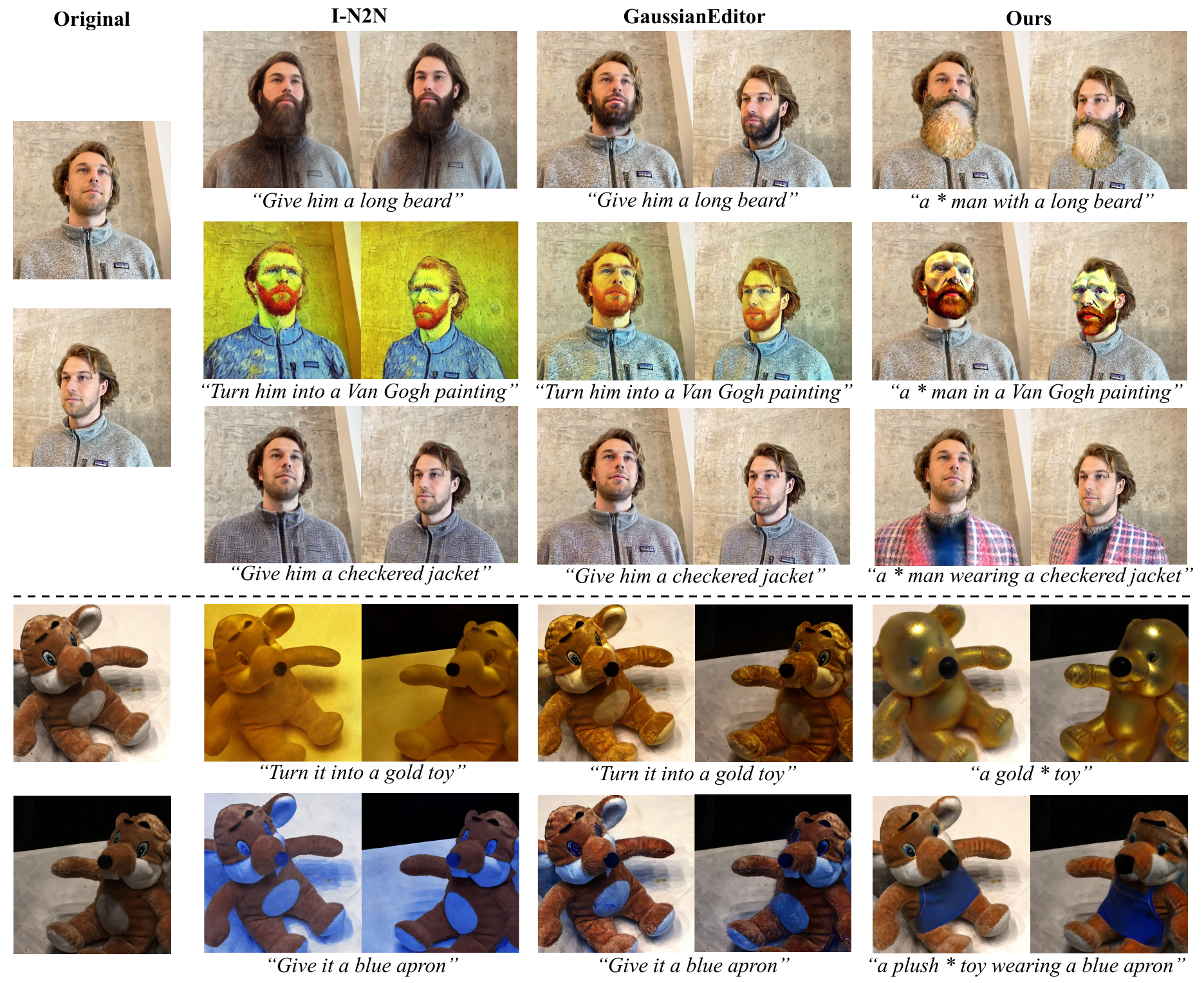}
  \caption{\label{fig:comp1}%
           \textbf{Comparisons with GaussianEditor\cite{chen2023gaussianeditor} and I-N2N\cite{haque2023instruct}.} We show superior ability in making successful edits and controlling editing regions, without leaking noisy Gaussians to backgrounds and irrelative regions.
           }
\end{figure*}

\begin{figure*}[tbp]
  \centering
\includegraphics[width=0.85\linewidth]{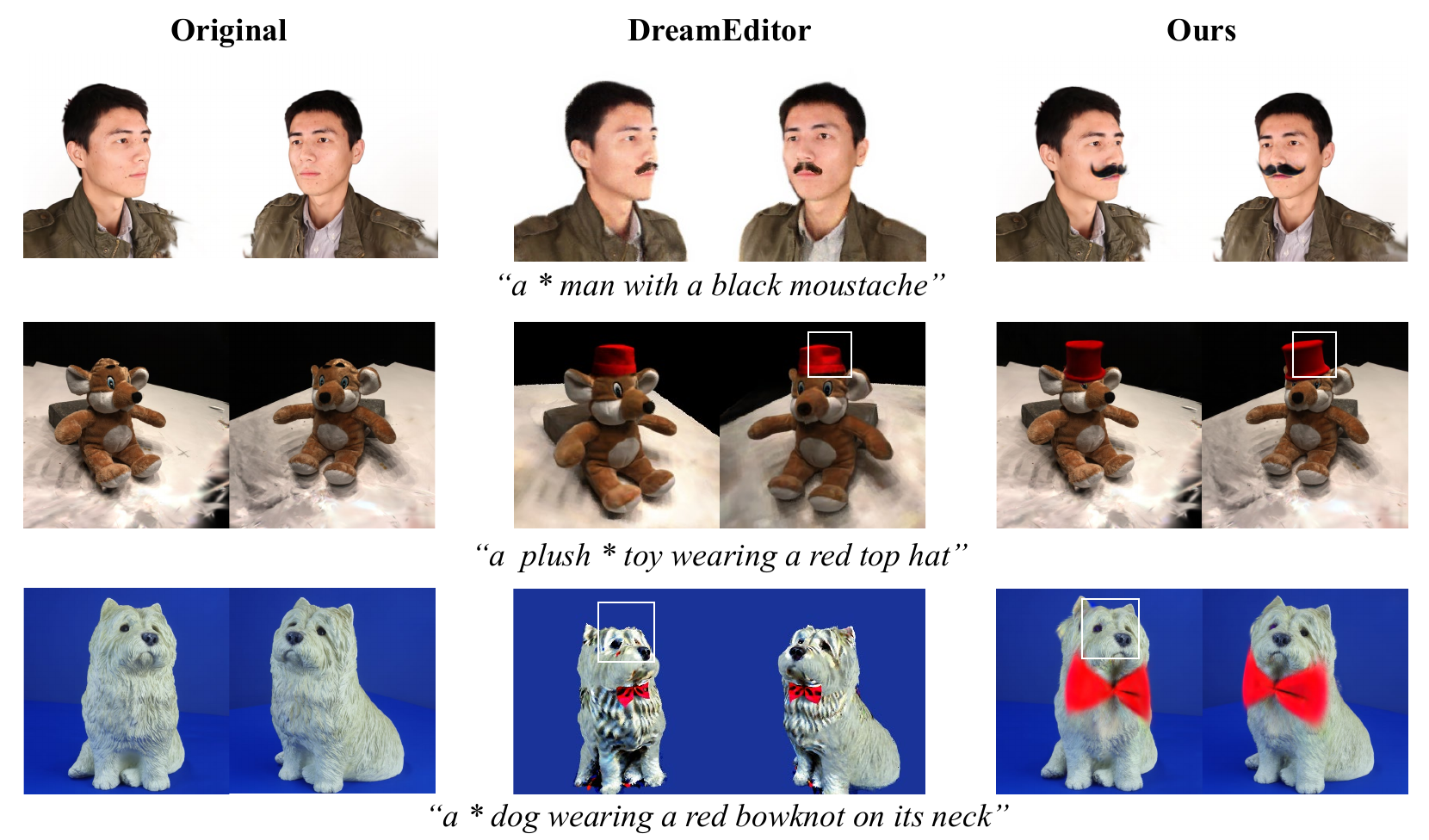}
  \caption{\label{fig:comp2}%
           \textbf{Comparisons with DreamEditor\cite{zhuang2023dreameditor}.} Our method adds a better-shaped mustache on the man, and a more detailed red top hat on the plush toy. For the dog, we make a more precise editing. Please be aware that the fragmented clothing in the first scene and the rough edges in the second scene result from the limited viewpoints used to train the original Gaussian models. This is an issue with the original Gaussians, but not with our editing method.
           }
\end{figure*}

\section{Experiments}

\subsection{Experimental Setup}
\label{sec:setup}
\textbf{Dataset.} 
We use scenes including human faces, indoor settings, and complex outdoor environments, providing a comprehensive evaluation of our method's efficacy. For outdoor environments, we use the Mip-NeRF360\cite{barron2022mip} dataset; for faces and indoor settings, we use datasets provided in GaussianEditor\cite{chen2023gaussianeditor} and DreamEditor\cite{zhuang2023dreameditor}. We employ the training method from 3D-GS\cite{kerbl20233d} and camera viewpoints selected from COLMAP\cite{schonberger2016structure} to train the original Gaussians. For each editing task, we use a text prompt about the scene as input and select a keyword to fine-tune the diffusion model.

\noindent
\textbf{Baseline.} We compare our approach with three baselines. Earlier text-based editing works often rely on NeRF\cite{mildenhall2021nerf}, so we select two representative text-based neural radiance field editing methods: Instruct-NeRF2NeRF\cite{haque2023instruct}(I-N2N) and DreamEditor\cite{zhuang2023dreameditor}. I-N2N utilizes Instruct-Pix2Pix\cite{brooks2023instructpix2pix} to update rendered multi-view images based on specific text instructions. DreamEditor employs a mesh-based representation and incorporates DreamBooth\cite{ruiz2023dreambooth} to support text-based editing. Additionally, for the latest editing works based on 3D-GS\cite{kerbl20233d}, we choose the state-of-the-art GaussianEditor\cite{chen2023gaussianeditor} for comparison. Similar to I-N2N, GaussianEditor utilizes Instruct-Pix2Pix to iteratively optimize parameters of Gaussians to guide 3D-GS in completing editing tasks.

\noindent
\textbf{Evaluation Criteria.} 
Following I-N2N and GaussianEditor, we use CLIP\cite{radford2021learning} text-image Directional Similarity (CLIP$_{dir}$). CLIP$_{dir}$ assesses the alignment between changes in text pairs and corresponding changes in image pairs. Detailed definitions will be presented in supplementary material. To ensure a fair comparison, we standardize prompts and instructions into the same format and generate the original text using the BLIP-2\cite{li2023blip2} model. 

Considering that the quality of editing is closely related to human perception, which cannot be fully quantified by CLIP$_{dir}$, we have also conducted user studies. We presented multi-view images of the original and edited scenes to participants and gathered their preferences. During the survey, we randomized the order of results so that users were unaware of which result belonged to which method.

\noindent
\textbf{Implementation Details.} 
For baseline comparisons, we mostly follow the recommended settings in their papers, except when extending the training iterations is necessary. 
For GaussianEditor and I-N2N, we make comparisons with a total of 14 editing tasks on 4 scenes. We collect 48 questionnaires for the user study.
For DreamEditor, considering that they have not published their preprocess method, we will conduct an additional comparison, using a subset of their released preprocessed datasets and checkpoints, with a total of 7 editing tasks on 3 scenes. We collect 56 questionnaires for the user study.

\subsection{Qualitative Results}
\label{sec:quali}
We provide qualitative results of our method in Figure~\ref{fig:teaser} and Figure~\ref{fig:more_result}. Results show that our method can properly locate editing regions and edit various scenes, which succeeds in altering the geometry and appearance of objects based on the text prompts.
Since GaussianEditor\cite{chen2023gaussianeditor} implements different pipelines for object insertion and other editing tasks, we will first make comparisons on addition with GaussianEditor and I-N2N\cite{haque2023instruct}. As shown in Figure~\ref{fig:comp-add}, we present results about adding sunglasses to the plush toy and a birthday hat to the man. GaussianEditor first selects a single view, repaints it, and makes monocular depth estimation, which cannot guarantee a precise depth map. So it tends to insert objects into the wrong locations. Besides, some unintended details can be observed(e.g. the face under the hat in the middle column of the first row). Although dynamically adjusting estimated depth can alleviate the first problem, the second one remains unsolved. To be supplemented, it is noticed that the details of the inserted object are not fine enough. I-N2N reveals its issue that it misunderstands prompts and generates confusing results, due to the limited ability of Instruct-Pix2Pix. In contrast, our method neither generates undesirable results nor mistakenly inserts objects into unexpected regions, leading to satisfactory and clean results.

As presented in Figure~\ref{fig:comp1}, we make comparisons on editing tasks. As mentioned before, I-N2N sometimes fails to handle complex instructions properly and cannot generate results corresponding to editing prompts (in the third row and the last row). Besides, it can be observed that editing regions spread all over the scene (in the second row and the fourth row) since I-N2N utilizes Instruct-Pix2Pix to fulfill dataset updates, which makes modifications to the whole picture. The second column shows the editing results of GaussianEditor, which demonstrates its limitations. In the first row, we try to generate a long beard for him, but the result doesn't meet our expectations. A possible reason is that its semantic labels cannot fully correspond with the prompt. However, our optimization guided by DreamBooth\cite{ruiz2023dreambooth} does not need to focus on this issue. Our pseudo-GT images prevent editing results from changing backgrounds (in the second row), while carefully designed guidance promises sufficient changes according to the prompt (in the  fourth row). Moreover, GaussianEditor inherits the shortcomings of I-N2N, as shown in the third and the last row, failing to edit correctly when instructions are complex.

Noticing that the editing pipelines in both GaussianEditor and I-N2N utilize Instruct-Pix2Pix, we also selected DreamEditor\cite{zhuang2023dreameditor} for another comparison. Results are shown in Figure~\ref{fig:comp2}. DreamEditor also fully leverages the capabilities of DreamBooth\cite{ruiz2023dreambooth}, achieving relatively high-quality edits on scenes. However, we can generate a beard with better shape (in the first row) and a hat with more details (in the second row). We also control the edit region, ensuring the dog does not appear noisy (in the last row). We believe that the key to our better results lies in the fact that driving Gaussians to the target region is easier than manipulating a mesh-based NeRF.
\subsection{Quantitative Results}
\label{sec:quanti}

We present the CLIP$_{dir}$ score and votes of users in Table~\ref{tab:t1} and Table~\ref{tab:t2}. In Table~\ref{tab:t1}, we compare our method with GaussianEditor\cite{chen2023gaussianeditor} and I-N2N\cite{haque2023instruct}. The results clearly indicate that our method achieves significantly higher CLIP$_{dir}$ scores and wider preference, suggesting better alignment between editing texts and results in our method. 
\begin{table}[htb]
    \centering
    \caption{\textbf{Quantitative comparison with GaussianEditor\cite{chen2023gaussianeditor} and I-N2N\cite{haque2023instruct} in CLIP Directional Similarity metrics and user study evaluations.}} \label{tab:t1}
    \begin{tabular}{ccc}
        \hline
        Method  & CLIP$_{dir}$$\uparrow$ & Vote$\uparrow$ \\
        \hline
        I-N2N  & 0.1542 & 23.1\% \\
        GaussianEditor & 0.1467 & 22.5\% \\
        Ours  & \textbf{0.1883} &  \textbf{54.4\%} \\
        \hline
    \end{tabular}
 \end{table}
 
In Table~\ref{tab:t2}, we compare with DreamEditor\cite{zhuang2023dreameditor}. The results indicate that our method achieves significantly higher CLIP$_{dir}$ scores, showing that the shapes and textures generated by our method are more consistent with the editing text prompts and with better quality. In conclusion, we found that the evaluation results further demonstrate that GSEditPro achieves higher user satisfaction in various scenarios, which also proves that the users prefer to more fantastic edit more than just texture edit.

\begin{table}[htb]
    \centering
    \caption{\textbf{Quantitative comparison with DreamEditor\cite{zhuang2023dreameditor} in CLIP Directional Similarity metrics and user study evaluations.}} \label{tab:t2}
    \begin{tabular}{ccc}
        \hline
        Method  & CLIP$_{dir}$$\uparrow$ & Vote$\uparrow$\\
        \hline
        DreamEditor  & 0.1911 & 38.1\% \ \\
        Ours  & \textbf{0.2021} &  \textbf{61.9\%}\ \\
        \hline
    \end{tabular}
 \end{table}

\begin{figure}[htb]
  \centering
  \includegraphics[width=1\linewidth]{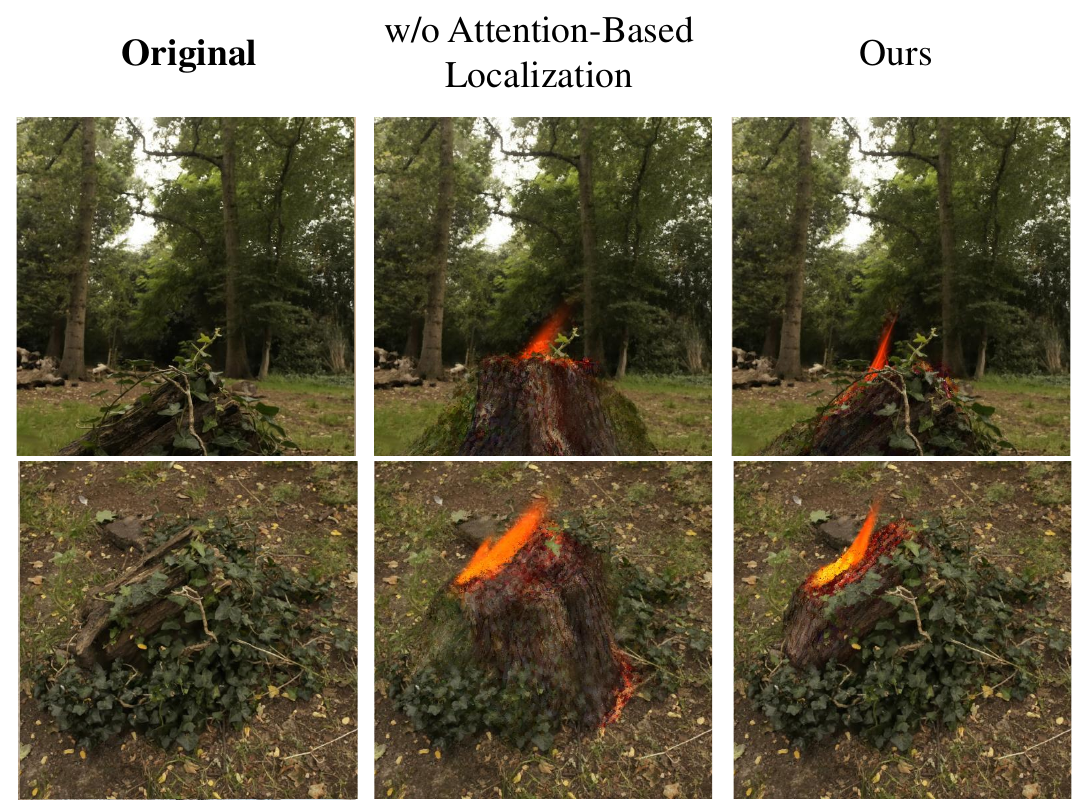}
  %
  %
  \caption{\label{fig:abla1}
           \textbf{Ablation study on attention-based localization of editing areas in 3D. Prompt: "a * stump on fire".} Utilizing attention-based localization in 3D allows for better preservation of the entire scene's details. Without our 3D localization, plants near the stump lose their form, whereas our complete method maintains the scene's similarity to the original scene from various perspectives.}
\end{figure}

\begin{figure}[htb]
  \centering
  \includegraphics[width=1\linewidth]{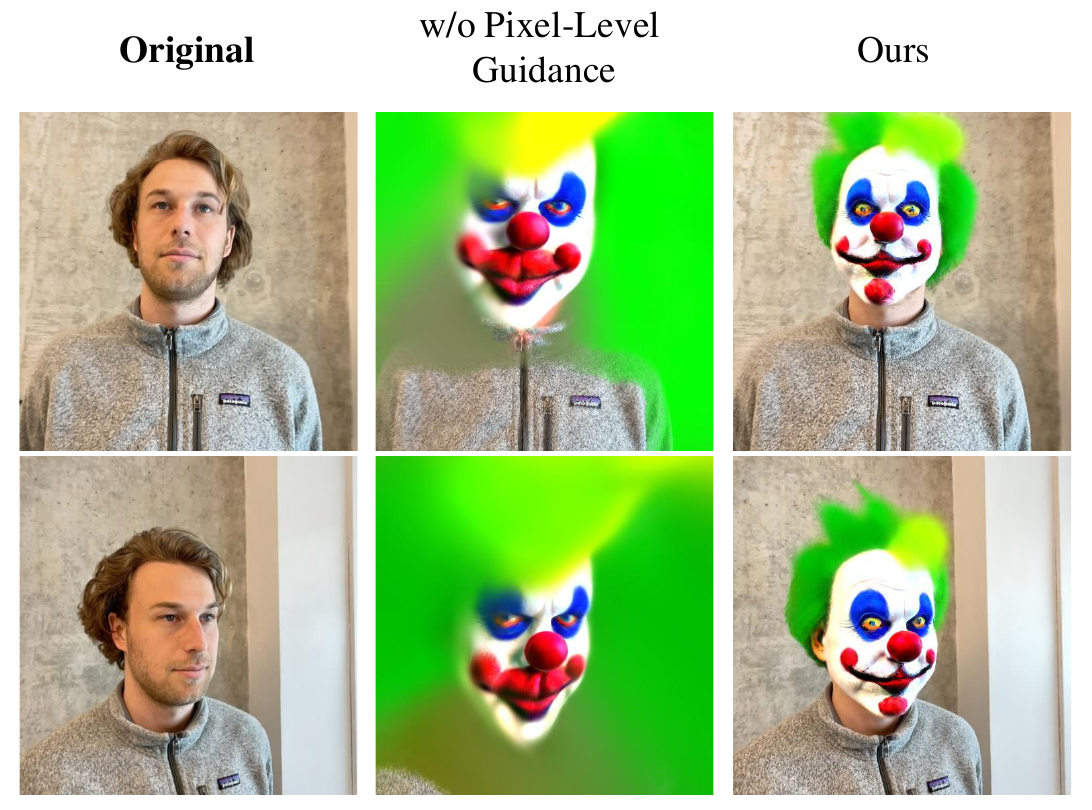}

  \caption{\label{fig:abla2}
           \textbf{Ablation study on Pixel-Level Guidance using pseudo-GT images. Prompt: "a * clown".} Due to the strong fluidity of Gaussians, the background remains cluttered with 3D static localization when pixel-level guidance is removed. However, with the guidance of pseudo-GT images, our method effectively controls the Gaussians, enabling detailed editing.}
\end{figure}

\subsection{Ablation Study}
\textbf{Effectiveness of our attention-based localization of editing areas.} To demonstrate the effectiveness of the precise localization of editing regions proposed in this paper, we conduct two experiments. (1) Without localization: We omit the localization step in Section~\ref{sec:locating} and directly optimize all Gaussians with the full steps afterward. (2) Our method: By adding semantic labels to Gaussian distributions, we precisely determine the editing region based on attention maps and only optimize the selected editing region during the optimization process. As shown in Figure~\ref{fig:abla1}, the method without localization inadvertently alters irrelevant areas in the scene(e.g. the plants near the stump), disrupting the consistency of non-editing areas. In contrast, with precise editing region localization, our method ensures that changes occur only in the regions of interest.

\noindent
\textbf{Effectiveness of the pixel-level guidance using pseudo-GT images}. Our detailed framework is accomplished by conducting score distillation sampling within the 3D editing mask region first and then further optimizing the quality with pseudo-GT images. Therefore, in ablation experiments, we design the following experiments. (1)Without pixel-level guidance: We remove the pixel-level guidance of pseudo-GT images and use the SDS guidance only. (2) Our method: Adding pseudo-GT images in training for further precise editing. As shown in Figure~\ref{fig:abla2}, it can be observed that in the results without pixel-level guidance, although we have located the editing regions in 3D as a constraint, which tries to enable the reduction of unnecessary modifications in irrelevant areas, the background is still messed up by the color from the hair of clown due the strong fluidity of Gaussians. However, when introducing pseudo-GT images for refinement, it can be visually observed from Figure~\ref{fig:abla2} that it effectively reduces artifacts of the background and better maintains the consistency of non-editing areas before and after editing, thereby greatly improving the editing quality.
\section{Conclusion}
We propose GSEditPro, a novel text-based 3D scene editing framework capable of performing various editing operations. By leveraging the explicit nature of 3D Gaussian distributions, we have devised a method to add semantic labels based on attention maps to each Gaussian during the differentiable rendering process, achieving precise localization of editing regions. Additionally, we have fine-tuned the diffusion model to optimize Gaussians and devised an attention-based method to preserve details through pixel-level guidance. Our extensive experiments have demonstrated that the attention-based progressive localization in both 2D and 3D significantly enhances our framework, outperforming previous methods even with less prior information.

\begin{figure}[htb]
  \centering
    \includegraphics[width=0.9\linewidth]{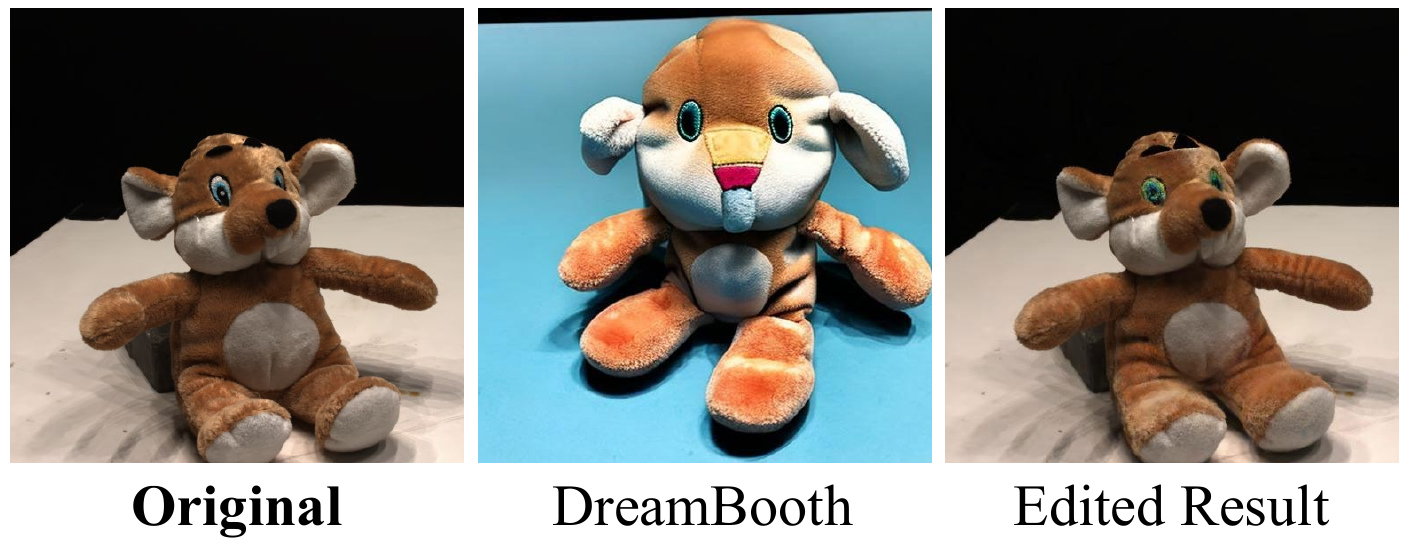}

\caption{
\label{fig:failcase}
           {\textbf{A failed case. Prompt: "a plush * toy wearing shoes". } 
           The middle column shows a bad generation of diffusion models, leading to the failure of editing(the right column).
           }}
\end{figure}

\noindent
\textbf{Limitation.} Our method heavily relies on the generation ability of the 2D diffusion models, which may cause our 3D editing to fail when the 2D generation from the diffusion models is poor. As shown in Figure~\ref{fig:failcase}, with the 2D supervision generated by the diffusion model (in the middle column), it will be difficult to make satisfactory editing results that match the prompt (in the right column).


\bibliographystyle{eg-alpha-doi} 
\bibliography{Main}       

\end{document}